\documentclass{article}

\usepackage{PRIMEarxiv}

\usepackage[utf8]{inputenc} 
\usepackage[T1]{fontenc}    
\usepackage{hyperref}       
\usepackage{url}            
\usepackage{booktabs}       
\usepackage{amsfonts}       
\usepackage{nicefrac}       
\usepackage{microtype}      
\usepackage{graphicx}       
\graphicspath{{media/}}     
\usepackage{multirow}       
\usepackage{natbib}

\title{Simplifying Scholarly Abstracts for Accessible Digital Libraries}

\author{
  Haining Wang \\
  Indiana University Bloomington\\
  Bloomington, Indiana, USA \\
  \texttt{hw56@indiana.edu} \\
  \And
  Jason Clark \\
  Montana State University \\
  Bozeman, Montana, USA \\
  \texttt{jaclark@montana.edu} \\
}

\begin{document}
\maketitle

\begin{abstract}
Standing at the forefront of knowledge dissemination, digital libraries curate vast collections of scientific literature. However, these scholarly writings are often laden with jargon and tailored for domain experts rather than the general public. As librarians, we strive to offer services to a diverse audience, including those with lower reading levels. To extend our services beyond mere access, we propose fine-tuning a language model to rewrite scholarly abstracts into more comprehensible versions, thereby making scholarly literature more accessible when requested. We began by introducing a corpus specifically designed for training models to simplify scholarly abstracts. This corpus consists of over three thousand pairs of abstracts and significance statements from diverse disciplines. We then fine-tuned four language models using this corpus. The outputs from the models were subsequently examined both quantitatively for accessibility and semantic coherence, and qualitatively for language quality, faithfulness, and completeness. Our findings show that the resulting models can improve readability by over three grade levels, while maintaining fidelity to the original content. Although commercial state-of-the-art models still hold an edge, our models are much more compact, can be deployed locally in an affordable manner, and alleviate the privacy concerns associated with using commercial models. We envision this work as a step toward more inclusive and accessible libraries, improving our services for young readers and those without a college degree.
\end{abstract}


\section{Introduction}
Making science more accessible remains a challenge even with much effort devoted on the producer and publisher side. As content producers, researchers are encouraged to engage directly with the public, either through social media \citep{davies2008constructing, hara2019emerging, knox2021public} or by crafting more digestible manuscripts in research \citep{maurer2021lessons} and practice \citep{grene2017use}. Funding agencies and renowned journals also encourage the communication of scientific findings in accessible language. For instance, the National Institutes of Health (NIH) advocate ``clear and simple'' principles when communicating with audiences with limited health literacy, and the \textit{Proceedings of the National Academy of Sciences of the United States of America} (\textit{PNAS}) requires authors to submit a significance statement accessible to non-experts \citep{berenbaum2021covid, pool2021infodemic}.

As scientific research progresses with increased specialization and interdisciplinarity, it is acknowledged that the use of jargon effectively reduces communication costs among domain experts, particularly those responsible for reviewing submissions. This specialized language, however, can become incomprehensible to those without a similar research background. While efforts to share scientific findings in more accessible language from the producer side are gaining traction, widespread adoption is unlikely in the near future due to the inherent conflicts between the specialized nature of scholarly communication and the public-oriented dissemination of scientific findings.

Within this effort to create understandable research findings and open science to broader communities, libraries—and our digital libraries in particular—have a role to play. Driven by this idea, we propose to start by improving the readability of abstracts from scholarly works through automated rewriting. Given the remarkable performance of language models, such as ChatGPT, across various tasks involving translation and summarization, we hypothesize that such models can perform well on a document simplification task: a translation-like task where the source is the original, possibly jargon-laden abstract, and the target is a more accessible version. Our ultimate goal is to serve these improved abstracts within digital library search results and indices to improve information retrieval, and most importantly, the understanding of the research itself.

Considering patron privacy and library budget, we propose fine-tuning a language model that can be deployed in-house. To this end, we introduced a novel corpus consisting of paired scientific abstracts and significance statements from diverse disciplines and fine-tuned four language models to test the idea of simplifying scholarly abstracts. By analyzing the outputs from the resulting models, we demonstrate their capacity to improve the readability of scholarly abstracts by roughly 3 points as measured using US grade-based readability scores, making a post-graduate abstract understandable by an undergraduate. We also qualitatively annotated the model outputs with respect to their language quality, faithfulness, and completeness. This revealed hidden patterns for each fine-tuned model, chief among them syntactic-level contributions to accessibility and emphasis on studies’ implications. Compared to the outputs of the most advanced commercial models, our findings indicate that there is room for word-level improvements in our models’ simplification. We envision this work as a step towards fulfilling libraries’ commitments to inclusivity and accessibility by bridging the gap in reference services for individuals with lower reading levels, such as young readers and those without a college degree.

Code, models, and their outputs are available \href{https://github.com/Wang-Haining/scholarly_abstract_simplification}{online} under permissive licenses.

\section{Related Work}
The community of natural language processing has a longstanding tradition of addressing document accessibility through automated rewriting, particularly via the task known as \textit{text simplification}. A common method is to fine-tune a language model with parallel corpora containing documents of different readability levels using cross-entropy objectives to ``translate'' complex texts into simpler versions. For instance, \citet{devaraj2021paragraph} fine-tuned a language model \citep{lewis2019bart} with a corpus of pairs of technical and plain-language medical texts. The fine-tuned model achieved about a 1-point Automated Readability Index (ARI) improvement in readability, on par with the readability level of the plain-language targets. Their best model achieved a 2-point ARI improvement by adding an extra term to the cross-entropy objective to reduce the probability assigned to a set of technical words identified using a separately trained classifier on essays of different readability levels \citep{xu2015problems}.

Efforts have also been made toward controllable text simplification, which finely controls the readability level of the generated document. For example, \citet{scarton2018learning} employed a neural translation model where the original sequence is prepended with tags indicating the target readability level, thereby controlling the generated text’s readability. \citet{nishihara2019controllable} further took word-level difficulty into account by adjusting the cross-entropy loss based on words frequently appearing in sentences of the target level. \citet{carlson2018evaluating} proposed evaluating the readability of simplified text using different versions of the Bible, which cover a wide spectrum of readability levels.

In short, it is promising to fine-tune a language model for more accessible language using a corpus that pairs abstracts with their pre-simplified versions.

\section{Scientific Abstract-Significance Statement (SASS) Corpus}
\label{sec: corpus_sass}

\begin{table}[!ht]
\caption{Corpus statistics for the Scientific Abstract-Significance Statement corpus. Metrics ARI, F-K, VOA, SL, WA, and WL stand for Automated Readability Index, Flesch-Kincaid readability test, log ratio of its proportion found in the VOA1500 vocabulary, average sentence length, number of sentences, word accessibility (i.e., log frequency per 1 billion tokens found in English Wikipedia), and average word length, respectively. Measures whose names are followed by a down arrow symbol ($\downarrow$) indicate that lower values correspond to a more readable document. Numeric values in parentheses are the corresponding standard deviations. Paired t-tests were conducted on each metric for the abstracts and significance statements, with p-values adjusted using the Bonferroni correction for multiple comparisons. The observed differences in each of the measurements are statistically significant after adjusting for the grouped p-values at a significance level of 0.05.}
\label{tbl: corpus_stats}
\small
\centering
\begin{tabular}{p{0.18\linewidth}p{0.06\linewidth}p{0.07\linewidth}p{0.07\linewidth}p{0.07\linewidth}p{0.07\linewidth}p{0.07\linewidth}p{0.07\linewidth}p{0.07\linewidth}}
\toprule
    Category & ARI$\downarrow$ & F-K$\downarrow$ & VOA & SL$\downarrow$ & WA & WL$\downarrow$   \\ \midrule
    Abstract & \scriptsize 18.9 \tiny (2.8) & \scriptsize 19.2 \tiny (2.4) & \scriptsize -0.43 \tiny (0.25) & \scriptsize 25.4 \tiny (4.9) & \scriptsize 12.0 \tiny (0.4) & \scriptsize 5.3 \tiny (0.4)  \\ \cmidrule{1-7}
    Significance statement & \scriptsize  18.1* \tiny (3.1)  & \scriptsize 18.6* \tiny (2.7) & \scriptsize -0.31* \tiny (0.26) & \scriptsize 23.9* \tiny (5.3)  & \scriptsize 11.9* \tiny (0.4) & \scriptsize 5.4* \tiny (0.4)  \\ 
\bottomrule
\end{tabular}
\end{table}

\begin{figure}[!ht]
  \centering
  \includegraphics[scale=0.45]{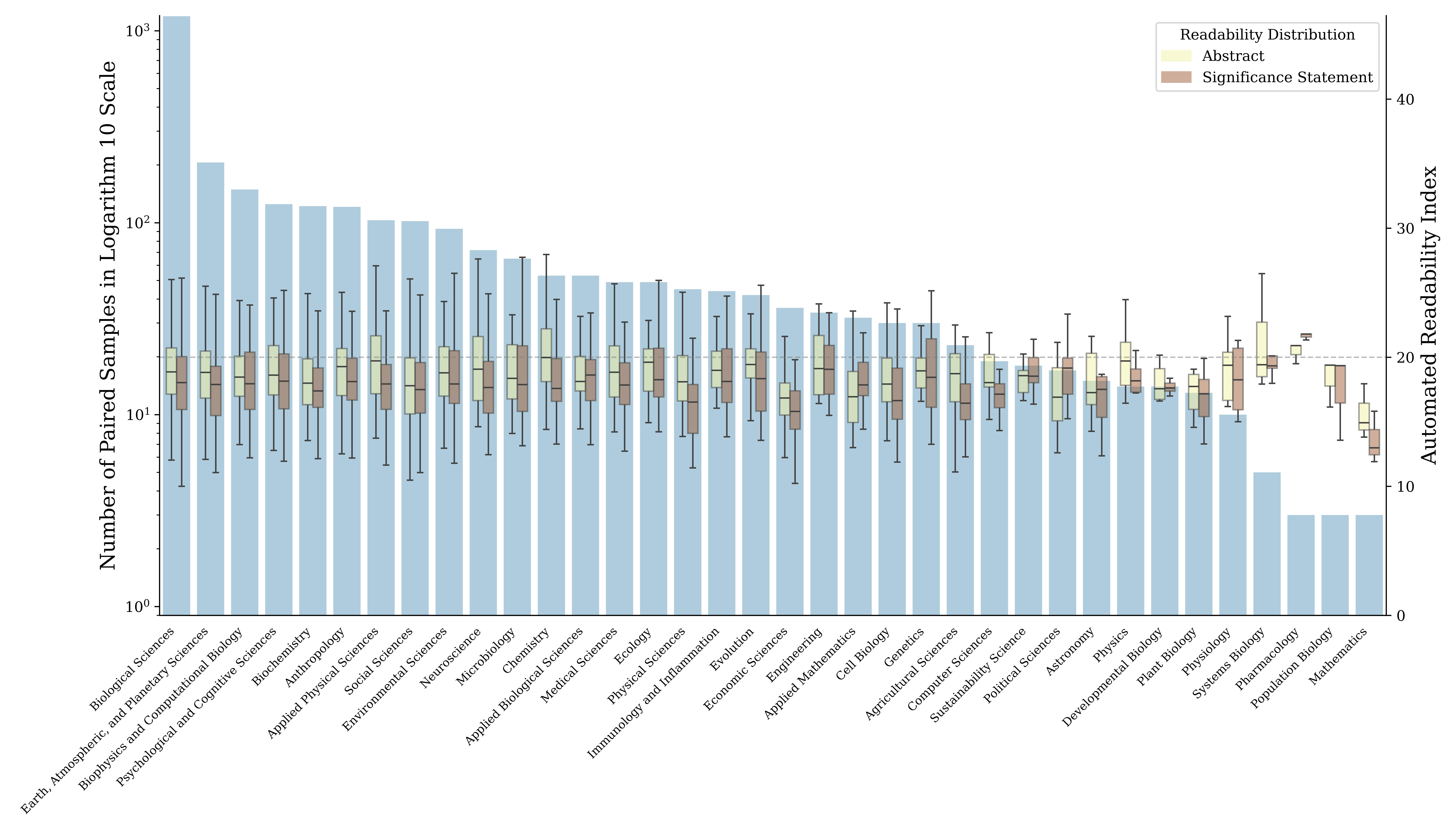}
  \caption{Discipline and readability distributions of abstracts and significance statements found in the training set of the Scientific Abstract-Significance Statement Corpus. The count of paired samples in different disciplines is shown in blue bars on a log10 scale (disciplines with fewer than three samples are not shown). Readability is measured using the Automated Readability Index (ARI), which estimates the number of years of schooling required to understand a text. On average, abstracts have a readability slightly below 20 ARI, indicating a post-graduate level. Significance statements are generally more readable than their corresponding abstracts.}
  \label{fig: discipline_dist}
\end{figure}

Recognizing the lack of a corpus suitable for scientific abstract simplification, we began by introducing a new dataset of paired texts composed of abstracts and significance statements derived from \textit{PNAS}. In October 2012, \textit{PNAS} started to require their authors to submit a statement of significance along with their manuscript that could be understood by ``an undergraduate-educated scientist outside their field of specialty'' \citep{verma2012pnas}. We observed that a majority of these statements are more readable than, yet semantically coherent with, their corresponding abstracts. The corpus covers a wide range of disciplines, ensuring diverse representation across various fields. Notably, there is a particularly high number of samples in biological sciences, followed by earth, atmospheric, and planetary sciences, and physical and social sciences, as shown in Figure~\ref{fig: discipline_dist}. We collected 3,430 abstract-significance statement pairs from cached \textit{PNAS} pages on the Internet Archive, cleaned them up, and divided them into training (3,030 samples), validation (200 samples), and test sets (200 samples). See corpus statistics in Table~\ref{tbl: corpus_stats}, and refer to Section~\ref{sec: evaluation} for the description of the measures.

The corpus statistics reveal that significance statements tend to be shorter and more readable than abstracts, as evidenced by lower mean values in the Automated Readability Index (ARI), Flesch-Kincaid readability test (F-K), and average sentence length metrics. This makes them useful for the task of simplifying scientific abstracts. Interestingly, word accessibility (i.e., log frequency per 1 billion tokens found in English Wikipedia) and average word length indicate that significance statements can be less accessible at the word level. Although the log ratio of words found in the VOA1500 vocabulary is slightly lower than in the corresponding abstracts, these 1,500 words are very basic and include a high proportion of function words. Considering that significance statements use approximately 1.5 fewer words on average, the increased use of VOA words may be a consequence of the higher use of function words to maintain grammaticality.

\section{Language Models \& Language Modeling}
Language models define a probability distribution over a sequence of words. For instance, given the sequence ``The cat sat on the,'' the model predicts the next token ``mat'' with a higher probability than ``the.'' The objective in training language models is to minimize prediction errors on the next word given the preceding words, using the cross-entropy loss function. Cross-entropy measures the difference between the predicted probability distribution $P$ and the true distribution $Q$. The formula for cross-entropy loss $H(P, Q)$ is:
\begin{equation}\label{eq: objective}
    H(P, Q) = -\sum_{i} Q(i) \log P(i)
\end{equation}
where $i$ represents a word in the vocabulary. Minimizing this loss trains the model to align its predictions closely with actual word occurrences.

The distribution of $Q$, at its core, encompasses all \textit{possible} discourse in the history of a language, whether spoken or written, natural or artificial, documented or not. In reality, only a fraction of human discourse is easily obtainable, primarily from the Internet, including sources such as textbooks, papers, social media, and transcriptions of movies and dialogues. Even so, the body of text used can be quite substantial: nowadays, it is not uncommon to train a language model with corpora consisting of trillions of words. This process of modeling such vast collections of human-produced documents is often referred to as \textit{pretraining}.

To make pretrained language models better at specific tasks, they are often fine-tuned using finely-crafted corpora suitable for tasks using the same objective function. For example, if we would like to use a pretrained language model for English-French translation, it is recommended to use concatenated English documents with their French equivalents, preferably with indicative words for the task such as ``Please translate this sentence to French: `The dog runs.'\,'' In the context of scholarly abstract simplification, we pair the abstracts with their significance statements during training via a template: ``Rewrite this abstract in plain English for middle school students: \verb|{abstract}|\verb|\n|Lay summary: \verb|{significance statement}|.'' During inference, providing an input without the \verb|{significance statement}| allows us to generate a document that follows the simplified language patterns learned from the training set, resulting in an accessible version.

\section{Experiment Setups}
\subsection{Language Models}\label{sec: lms}
We used four causal language models, OLMo-1B, Gemma-2B/-7B, and Phi-2 (2.7B), as the base models for separate experiments. OLMo-1B (Open Language Model) has 1.18 billion parameters. It is dedicated to improving language modeling studies with transparency \citep{groeneveld2024olmo} and developed by the Allen Institute for AI (AI2). OLMo-1B is pretrained with 2.46 trillion tokens on the Dolma dataset, an open corpus consisting of three trillion tokens derived from a mix of web content, scholarly publications, code, books, social media, and encyclopedic materials \citep{soldaini2024dolma}. Gemma-2B and Gemma-7B are developed by Google and trained on three trillion and six trillion tokens respectively, consisting of publicly available data as well as proprietary datasets comprising ``primarily-English data from web documents, mathematics, and code'' \citep{gemmateam2024gemma}. Phi-2 has 2.78 billion parameters and has been exposed to 1.4 trillion tokens, including subsets of Python code from The Stack v1.2, Q\&A content from StackOverflow, competition code from code\_contests, and synthetic Python textbooks and exercises generated by GPT-3.5 \citep{javaheripi2023phi}.

The models differ in their training details, particularly regarding the training corpus and size, as well as technical aspects such as activation functions and context length. They were chosen for their proven performance and relatively compact sizes.

\subsection{Training}  
We fine-tuned each of the language models using a cross-entropy objective (Eq.~\ref{eq: objective}), maximizing the likelihood of predicting the next word based on previous words in the training set of the SASS corpus. Each abstract was prepended with the same instruction and appended with an indicator marking the start of the corresponding significance statement. The learning rate for the Gemma-2B and Phi-2 experiments was set to $1 \times 10^{-5}$, for OLMo-1B it was set to $3 \times 10^{-6}$, and for Gemma-7B it was set to $1 \times 10^{-6}$. All experiments used a weight decay of $0.1$ and $50$ warm-up steps. The optimization used standard AdamW optimizer parameters ($\beta_1 = 0.9$, $\beta_2 = 0.999$, $\epsilon = 1 \times 10^{-8}$) \citep{kingma2014adam, loshchilov2017decoupled}.

We trained each language model three times with different random seeds and reported the performance of the supervised fine-tuned models using the checkpoints with the optimal validation loss. The batch size was set to 4 for OLMo-1B and Gemma-2B, and 2 for Phi-2 and Gemma-7B. Gradient accumulation steps were set to 4. Training was conducted on an A40 (48GB memory) GPU for smaller models and an H100 for Gemma-7B using ZeRO stage 2 CPU offloading \citep{rajbhandari2020zero}.

\subsection{Zero-Shot Performance of GPT-3.5 and GPT-4o}
We also measured the performance of state-of-the-art commercial models, i.e., GPT-3.5 (\verb|gpt-3.5-turbo-0125|) and GPT-4o (\verb|gpt-4o-2024-05-13|), on the test set of the SASS corpus. The system prompt was minimalist, designed to output in a flat JSON format for ease of extracting the accessible version.\footnote{The system instruction reads: `You are a helpful assistant designed to output JSON with a single key ``simplified\_version''. Ensure there is no nesting in the JSON structure.'} The instruction and sampling method are the same as those used in our models.

Since we did not explicitly expose our training set or provide extra demonstrations, directly instructing a model is termed zero-shot learning \citep{radford2019language}.\footnote{We cannot be certain that our corpus was not exposed to GPT-3.5/4o in their colossal yet closed-source datasets. Given that \textit{PNAS} hosts manuscripts online, we cannot deny the probability.} The major difference between supervised fine-tuning and zero-shot learning is that, in the former, the model weights are updated according to the objective (Eq.~\ref{eq: objective}). In contrast, in zero-shot learning, the model’s weights remain unchanged, as it has already been trained to function as a helpful and harmless chatbot \citep{ouyang2022training, chung2024scaling}. The performance of these models provides an upper bound estimate for how well a much lighter and more open model such as ours can optimistically perform in the foreseeable future.

\subsection{Evaluation}
\label{sec: evaluation}
We evaluated the outputs generated by the fine-tuned language models using the test set from the SASS corpus, which comprises 200 abstracts to be simplified. While we acknowledge that advanced decoding methods might refine the quality of the outputs, we used multinomial sampling with the temperature set to zero in this study to intentionally produce the most deterministic outputs, aiming to understand their modeling of accessible language and identify potential quirks.

The quality of the generated simplified abstracts is assessed both quantitatively and qualitatively. Quantitatively, we measure the outputs based on their semantic retention and accessibility using established relevance measures and metrics for readability and accessibility.

\subsubsection{Semantic Retention}
BERTScore calculates the cosine similarity for each token in the candidate sentence with each token in the gold reference in the space of the pre-trained contextual embeddings from a language model; its results align well with human judgment on semantic similarity evaluation. It is not influenced by lexical overlap, making it more suitable for evaluating simplification systems. For our evaluation, we used embeddings from the 18th layer of a BERT-large-uncased model and reported the F1 score. This choice is based on prior findings indicating that the 18th layer yields a high Pearson correlation (0.72) on the WMT16 To-English benchmark \citep{zhang2019bertscore}.

\subsubsection{Accessibility}
Accessibility can be measured with respect to the overall simplification quality (SARI); readability (ARI and Flesch-Kincaid); and other straightforward document complexity measures, including average sentence length, word accessibility (i.e., log frequency per 1 billion tokens found in English Wikipedia), the log ratio of its proportion of VOA Special English words (1,517 types in total), and average word length.

SARI (System output Against References and against the Input sentence) is specifically designed to evaluate text simplification \citep{xu2016sari}. It aims to measure how well a simplified text retains the original meaning while improving readability. SARI provides a balanced measure of how well a text simplification system performs by focusing on the necessary operations of adding, deleting, and retaining words.

ARI and Flesch-Kincaid readability tests assign a numerical score to text that reflects the U.S. educational grade level necessary for understanding. Lower scores (1--13) indicate content suitable for kindergarten through twelfth grade, with each score corresponding to a subsequent grade level. Scores in the range of 14--18 suggest college-level readability, ranging from first- to senior-year content appropriateness. Higher scores (19 and above) are associated with advanced college education. Both measures use average sentence length. Flesch-Kincaid uses syllables per word, while ARI uses characters per word for its linear combination with sentence length.

We harvested VOA Special English vocabulary comprising 1,517 unique words (VOA1500).\footnote{We included VOA Special English Word Book Sections A-Z, Science programs, and Organs of the body hosted on Wikipedia (\url{https://simple.wikipedia.org/wiki/Wikipedia:VOA_Special_English_Word_Book}).} We count the ratio of the words that appear in VOA1500 versus those that do not, and report the natural logarithm of the proportion of VOA words found in a generated sample. Values above 0 indicate that the text contains more Special English words than non-Special English words, and a higher value indicates a greater presence of easy words.

\subsubsection{Language Quality, Faithfulness, \& Completeness}
We manually examined 5\% of all generated samples, corresponding to a randomly chosen subset of the test set from the SASS corpus. Each generation is annotated with respect to language quality, faithfulness, and completeness using a rubric of Good, Acceptable, and Poor. We focused on fluency and grammaticality and hand-picked both good and problematic examples when evaluating language quality. Admittedly, there is no effective way to completely eliminate hallucinations in language models, but our abstract simplification system should not produce misinformation. Faithfulness to the original content, especially for the latest research where authentic sources are not available, lacks a well-acknowledged automatic measure. Lastly, it is essential to include the main findings and implications of the research for the general public, as this is the goal of scientific dissemination.

\begin{table}[!ht]
\caption{Comparison of performance from language models fine-tuned with the training set of the Scientific Abstract-Significance Statement corpus. Inference on the test split from SASS uses multinomial sampling at temperature zero. Metrics ARI, F-K, SARI, VOA, SL, WA, WL, and BS stand for Automated Readability Index, Flesch-Kincaid readability test, log ratio of VOA1500 vocabulary, sentence length, word accessibility, word length, and BERTScore (F1), respectively. Measures followed by a down arrow symbol ($\downarrow$) indicate that lower values are better. Numeric values in parentheses are the corresponding standard deviations. A paired two-tailed t-test was performed on observations of each measure between each model and the original abstracts, except for BS and SARI. At a model-wise p-value of 0.05, all measures differ significantly from the original abstracts.}
\label{tbl: automatic_evaluation}
 \small
 \centering
 \begin{tabular}{p{0.10\linewidth}p{0.08\linewidth}p{0.07\linewidth}p{0.07\linewidth}p{0.07\linewidth}p{0.07\linewidth}p{0.06\linewidth}p{0.06\linewidth}p{0.06\linewidth}p{0.06\linewidth}}
    \toprule
        Model & Setup  & ARI$\downarrow$ & F-K$\downarrow$ & SARI & VOA & SL$\downarrow$ & WA & WL$\downarrow$ & BS  \\ \midrule
        OLMo-1B & SFT    & \scriptsize 16.1 \tiny (3.3) & \scriptsize 17.0 \tiny (2.8) & \scriptsize 37.6 \tiny (4.6) & \scriptsize -0.29 \tiny (0.35) & \scriptsize 22.2 \tiny (5.4)  & \scriptsize 11.9 \tiny (0.5) & \scriptsize 5.1 \tiny (0.4) & \scriptsize 0.63 \tiny (0.06)\\  
        \cmidrule{1-10}
        Gemma-2B & SFT   & \scriptsize 15.5 \tiny (3.0) & \scriptsize 16.5 \tiny (2.6) & \scriptsize 39.1 \tiny (5.0) & \scriptsize -0.26 \tiny (0.30) & \scriptsize 20.6 \tiny (4.1)  & \scriptsize 11.9 \tiny (0.5) & \scriptsize 5.2 \tiny (0.4) & \scriptsize 0.64 \tiny (0.06)\\  
        \cmidrule{1-10}
        Phi-2 (2.7B) & SFT   & \scriptsize 12.5 \tiny (7.5) & \scriptsize 14.3 \tiny (6.2) & \scriptsize 37.7 \tiny (5.4) & \scriptsize -0.03 \tiny (0.46) & \scriptsize 21.1 \tiny (12.7)  & \scriptsize 12.5 \tiny (0.7) & \scriptsize 4.5 \tiny (0.6) & \scriptsize 0.53 \tiny (0.08)\\  
        \cmidrule{1-10}
        Gemma-7B & SFT   & \scriptsize 16.3 \tiny (2.6) & \scriptsize 17.1 \tiny (2.2) & \scriptsize 43.7 \tiny (9.8) & \scriptsize -0.27 \tiny (0.25) & \scriptsize 21.9 \tiny (3.9)  & \scriptsize 12.0 \tiny (0.4) & \scriptsize 5.2 \tiny (0.4) & \scriptsize 0.67 \tiny (0.07)\\  
        \cmidrule{1-10}
        GPT-3.5 & Zero-shot  & \scriptsize 9.4 \tiny (2.0) & \scriptsize 9.7 \tiny (1.7)  & \scriptsize 36.8 \tiny (4.5) & \scriptsize 0.27 \tiny (0.26) & \scriptsize 16.8 \tiny (3.0)  & \scriptsize 12.5 \tiny (0.3) & \scriptsize 4.4 \tiny (0.3) & \scriptsize 0.58 \tiny (0.04) \\ \cmidrule{1-10}
        GPT-4o & Zero-shot  & \scriptsize 8.5 \tiny (1.7) & \scriptsize 8.9 \tiny (1.7) & \scriptsize 37.5 \tiny (4.6) & \scriptsize 0.10 \tiny (0.25) & \scriptsize 14.5 \tiny (2.3)  & \scriptsize 12.2 \tiny (0.3) & \scriptsize 4.4 \tiny (0.3) & \scriptsize 0.59 \tiny (0.04)\\ 
    \bottomrule
  \end{tabular}
\end{table}

\section{Findings}
\label{sec: findings}
\subsection{Quantitative Assessment}
Table~\ref{tbl: automatic_evaluation} summarizes the performance of OLMo-1B, Gemma-2B/-7B, and Phi-2 (2.7B) when simplifying abstracts in the SASS corpus test set after tuning on its training set. We assessed the generation quality by considering both semantic retention (using BERTScore; BS for short), and accessibility, via ARI, Flesch-Kincaid (F-K), SARI, log ratio of VOA1500 vocabulary (VOA), sentence length (SL), word accessibility (WA), and word length (WL). A paired two-tailed related t-test was conducted on observations of each measure between the model outputs and the corresponding original abstracts, except for BERTScore and SARI. Bonferroni correction was applied to maintain a model-wise false discovery rate at 0.05.

The outputs from the three models fine-tuned with our SASS corpus show that all measures indicate significant improvements compared to the original abstracts. Specifically, the ARI and Flesch-Kincaid readability tests demonstrate a significant increase in readability, with an average improvement of over 3 points in ARI. These improvements suggest that the systems have the potential to reduce professional-level scholarly abstracts to a level comprehensible to undergraduate readers. The performance of OLMo-1B and Gemma-2B/-7B is consistent across the test set, which consists of abstracts from diverse disciplines. Although Phi-2 holds the lowest readability scores, we observed higher-than-expected variance in readability measures, suggesting inconsistencies in its output quality. After manual examination, we found that it performs reasonably for some test samples but fails for others by repeating simple yet meaningless phrases. Therefore, we reported model performance focusing on OLMo-1B and Gemma-2B/-7B, and will discuss Phi-2's behavior in Section~\ref{sec: discussion}. In zero-shot scenarios, GPT-3.5 and GPT-4o can produce simplified abstracts with a readability around 9 ARI, corresponding to a middle school student level.

We observed that the improvement in readability in our models results mostly from syntactic modification and marginally from word-level changes. Notably, sentence length was reduced by ca. 4 words on average, which has a significant influence on the calculation of ARI and Flesch-Kincaid scores; however, only about 0.2--0.3 characters were reduced at the word level, in contrast to GPT-3.5 and GPT-4o's almost 1-character reduction. The improvements in the difficulty level of individual words are not obvious: this is evidenced by both the VOA1500 vocabulary log ratio and the word accessibility (log frequency per 1 billion words found in English Wikipedia) for OLMo-1B and Gemma-2B/-7B. Although statistically different from the original abstracts, the effect should be considered marginal given the relatively large standard deviation.

The SARI score reflects a simplification system’s ability to retain essential content while making necessary additions and deletions to improve readability. We observe that the SARI scores for the fine-tuned models perform on par with GPT-3.5 and GPT-4o in zero-shot scenarios. Since SARI measures three abilities in balance, an intuitive interpretation of the score on its own is not readily available. The BERTScore shows that the substantive semantics of the abstracts are well-preserved post-simplification. Considering this information for the interpretation of SARI, the similar SARI scores may be due to the fact that our models retain more information from the original abstracts: in fact, over 5\% more than do GPT-3.5 and GPT-4o as measured in the latent space. Overall, these figures indicate that OLMo-1B and Gemma-2B/-7B can rewrite abstracts into more readable versions while retaining substantive semantics.

\subsection{Qualitative Assessment}

\begin{figure}[!ht]
  \centering
  \includegraphics[scale=0.37]{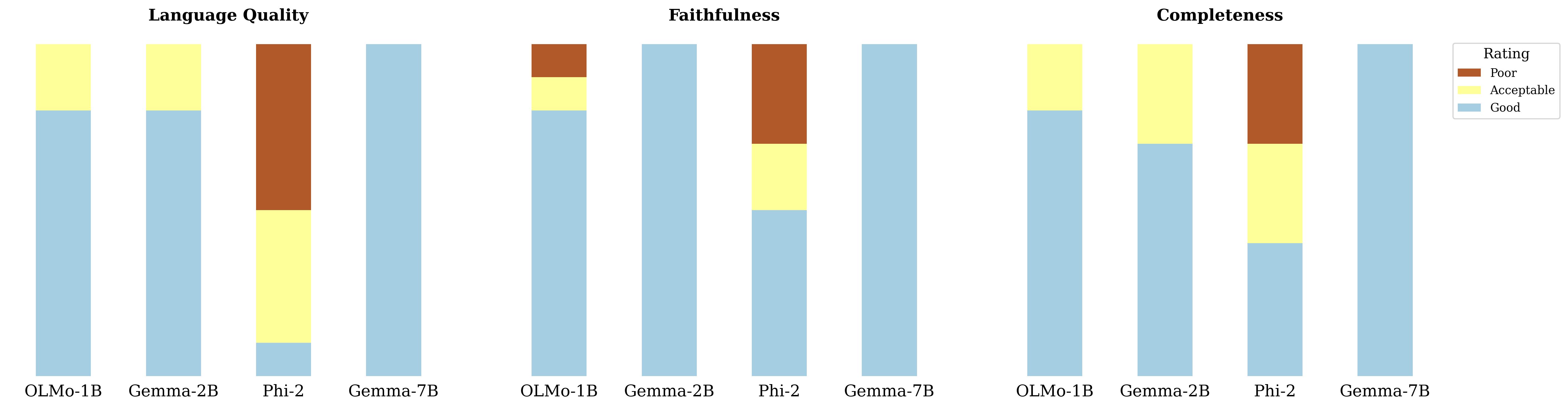}
  \caption{We annotated 5\% of generated outputs from OLMo-1B, Gemma-2B/-7B, and Phi-2 with respect to language quality, faithfulness, and completeness. The fine-tuned Gemma-7B performed the best on balance, followed by Gemma-2B and OLMo-1B.}
  \label{fig: annotation}
\end{figure}

We annotated 5\% of outputs generated from the fine-tuned models with respect to language quality, faithfulness, and completeness, as shown in Figure~\ref{fig: annotation}. The samples included three from biological sciences, and one each from chemistry, mathematics, evolution, environmental sciences, ecology, economic sciences, and earth, atmospheric, and planetary sciences.

We did not find any language issues in the outputs from Gemma-7B and found that the fine-tuned OLMo-1B and Gemma-2B produced overall high-quality language. However, Phi-2 generally fell short of our expectations. Interestingly, Gemma-2B produced two outputs that verbatim copied the input: one was an overly short abstract in mathematics, and the other had a readability score of approximately 16, which is only slightly higher than the readability of its average generation. Gemma-2B also occasionally ended with strange trailing notes, such as ``(PsycInfo Database Record (c) 2021 APA, all rights reserved),'' which does not appear in our training set. We hypothesize that this is carried over from the pretraining phase. Phi-2 often produced sensible words initially, but then developed into repetitions of short, accessible versions, one-sentence summaries, or meaningless phrases. We hypothesize that the problematic outputs were worsened by our naive sampling strategy, making it difficult to assert that Phi-2 is readily applicable in practice.

In assessments of faithfulness, we found almost perfect output faithfulness in Gemma-2B/-7B, although the former avoided the most challenging, short abstract by outputting it verbatim. While this is the least harmful strategy a language model can perform when the input abstract is unclear, we found hallucinations of a coined virus name and a not immediately relevant claim from OLMo-1B, along with two other marginal deviations, such as a coherent accessible summary followed by ``The first thing you need to [...]'' This sentence does not appear to be generated by random chance, as we observed such behavior in other generations as well as in OLMo-1B trained with different random seeds. Both Gemma-2B and OLMo-1B reduced the technical parts of the original abstract and elaborated on research implications. Repetitions from Phi-2 diluted the content, rendering half of the outputs only marginally relevant.

OLMo-1B and Gemma-2B often exhibit an inclination to over-emphasize implications, frequently leading to unfinished sentences within the constraints of the token budget we imposed for newly generated content. In comparison, Gemma-7B does not tend to overstate implications or repeat information from previous sentences. These outputs are classified as having acceptable completeness if the main implication is already clear. This issue may be alleviated with carefully-crafted, model-specific decoding strategies, such as nucleus sampling \citep{holtzman2019curious} or repetition penalty \citep{keskar2019ctrl}.

Word-level analysis reveals the main drawback of the language models during simplification: the lack of substitution of jargon with more accessible words. In almost all cases, the models either copy the original technical terms, avoid mentioning them in the newly composed version, or, worse, invent new terms. Consequently, the improvements in the ratio of VOA1500 vocabulary and word accessibility in quantitative assessments originate primarily from syntactic changes.

\section{Discussion and Further Research}
\label{sec: discussion}
To better serve individuals with lower reading levels, we fine-tuned OLMo-1B, Gemma-2B/-7B, and Phi-2 using a new parallel corpus consisting of accessible abstracts from diverse disciplines. We found that the fine-tuned OLMo-1B and Gemma-2B/-7B models can produce more accessible versions of a given abstract while retaining substantive semantics in high-quality language. These models should be considered solid baselines for further improvements, as we did not tune hyperparameters nor use advanced decoding strategies.

We investigated the quality of the generated outputs and concluded that the improvements in readability are mostly due to shorter sentences, while word-level improvements are not notable. Except for Gemma-7B, the language models are overly prone to emphasizing the significance and implications of research. Both of these tendencies may be attributed to the inherent challenges in balancing technical precision with accessibility in scholarly writing and the nature of significance statements, where authors aim to convey the importance of their research without compromising academic rigor. From the perspective of the training objective, this is expected since we did not find a notably higher use of accessible words in the significance statements of the SASS corpus.

We also caught Gemma-2B’s ``benign cheating,'' in the form of occasionally appending meaningless trailing phrases, when the input was already accessible enough. In an interesting comparison between the Gemma models, Gemma-7B effectively avoids repetition and stopping at the right point. Given the same naive decoding strategy, we believe Gemma-7B better captures the characteristics of the distribution of the significance statements. These improvements may be attributed to its tripled model size and doubled training data. For Phi-2, we hypothesize that its unstable performance and repetition behavior are partly due to its over-reliance on GPT-3.5/4 for pretraining. Phi-2 was trained using synthetic data created by GPT-3.5 and filtered web data using GPT-4, in which case the larger models’ biases are likely passed to the ``distilled'' model \citep{yu2024large}.\footnote{We invested additional time in fine-tuning Phi-2, but the resulting models performed similarly.}

Our findings largely corroborate previous research, indicating that language models fine-tuned on relevant parallel corpora can produce documents with readability levels comparable to, or slightly lower than, their plain-language counterparts \citep{devaraj2021paragraph, xu2015problems}. This outcome is expected, as the objective of supervised fine-tuning is to minimize the Kullback–Leibler divergence between the token distribution of the target readability levels and that of our language models. Consequently, the best result achievable through supervised fine-tuning is to produce lay versions that conform to the target distribution.

The generated versions, however, average an ARI of 15 to 16, which is significantly more readable than the target distribution of human-written significance statements with an average ARI of approximately 18. We hypothesize that this extra performance gain is primarily due to the extensive pretraining corpus to which our models were exposed. Additionally, this may be partly attributed to the differences between model architectures and training objectives, as we are using causal language models (i.e., decoder-only architecture \citep{radford2018improving}), while others use classic Transformer blocks (i.e., encoder-decoder architecture \citep{vaswani2017attention}) trained with a masked language model objective. Although there is still debate regarding the superior performance of one architecture over the other, causal language models are the mainstream choice due to their simple training objective and long context.

Although state-of-the-art commercial models still hold an edge, our fine-tuned language models are more affordable, considering typically tight budgets, and proactively address patrons’ privacy concerns by working in-house. Additionally, we foresee that the generated abstracts can enhance information retrieval by incorporating them into the search index of a digital library and delivering them to relevant parties through the user interface. Future steps include guiding the models via reinforcement learning, using rewards with an emphasis on improving word-level accessibility.

\section*{Acknowledgment}
We are grateful for the support from the Institute of Museum and Library Services (IMLS) under grant RE-246450-OLS-20. 
Computational tasks were carried out on the Tempest High Performance Computing System, which is operated and supported by the University Information Technology Research Cyberinfrastructure at Montana State University.

\bibliographystyle{acl_natbib}  
\bibliography{references}  

\end{document}